%% file: constantSGD.tex
\documentclass[]{article}
\usepackage[accepted]{icml2016}
\usepackage{fancyhdr}

\usepackage{bbm}
\usepackage{color}
\newcommand{\be}{\begin{eqnarray}}
\newcommand{\ee}{\end{eqnarray}}
\newcommand{\E}{\mathbb{E}}
\newcommand{\n}{\nonumber \\}

\usepackage{natbib}
\usepackage{url}
\usepackage{graphicx}
\usepackage{amsmath,amsthm,amssymb,graphicx, algorithm,subfig,float,times}
\usepackage{txfonts}
 \usepackage{enumitem}

\graphicspath{{../plt/}{../plt/confirm_theory/}}

\definecolor{darkmidnightblue}{rgb}{0.0, 0.2, 0.4}
\definecolor{denim}{rgb}{0.08, 0.38, 0.74}
\definecolor{royalblue}{rgb}{0.0, 0.14, 0.4}
\usepackage[colorlinks=true]{hyperref}
\hypersetup{citecolor=royalblue}
\hypersetup{linkcolor=royalblue}

\newcommand{\btheta}{\mathbf{\theta}}
\newcommand{\bx}{\mathbf{x}}
\newcommand{\g}{\, | \,}
\newcommand{\parhead}[1]{\textbf{#1 \,}}

\begin{document}

\twocolumn[
\icmltitle{A Variational Analysis of Stochastic Gradient Algorithms}

\icmlauthor{Stephan Mandt}{sm3976@columbia.edu}
\icmladdress{Columbia University}
\icmlauthor{Matthew D. Hoffman}{mathoffm@adobe.com}
\icmladdress{Adobe Research}
\icmlauthor{David M. Blei}{david.blei@columbia.edu}
\icmladdress{Columbia University}

\icmlkeywords{machine learning}

\vskip 0.3in
]

\date{\today}

\begin{abstract}
  Stochastic Gradient Descent (SGD) is an important algorithm in
  machine learning.  With constant learning rates, it is a stochastic
  process that, after an initial phase of convergence, generates
  samples from a stationary distribution.  We show that SGD with
  constant rates can be effectively used as an approximate posterior
  inference algorithm for probabilistic modeling.  Specifically, we show
  how to adjust the tuning parameters of SGD such as to match the
  resulting stationary distribution to the posterior.  This analysis
  rests on interpreting SGD as a continuous-time stochastic process
  and then minimizing the Kullback-Leibler divergence between its
  stationary distribution and the target posterior.  (This is in the
  spirit of variational inference.)  In more detail, we model SGD as a
  multivariate Ornstein-Uhlenbeck process and then use properties of
  this process to derive the optimal parameters.  This theoretical framework also connects
  SGD to modern scalable inference algorithms; we analyze the recently
  proposed stochastic gradient Fisher scoring under this perspective.
  We demonstrate that SGD with properly chosen constant rates gives a new way to
  optimize hyperparameters in probabilistic models.

\end{abstract}

\input{introduction}

\input{framework}

\input{inference.tex}

\section{Experiments}

We test our theoretical assumptions in section~\ref{sec:toy_data} and find
good experimental evidence that they are correct. In this section, we compare
against other approximate inference algorithms. In section~\ref{sec:hyper_exp} we show that 
constant SGD lets us optimize hyperparameters in a Bayesian model.

\label{sec:experiments}

\begin{figure}
\begin{center}
\includegraphics[width=0.48\linewidth]{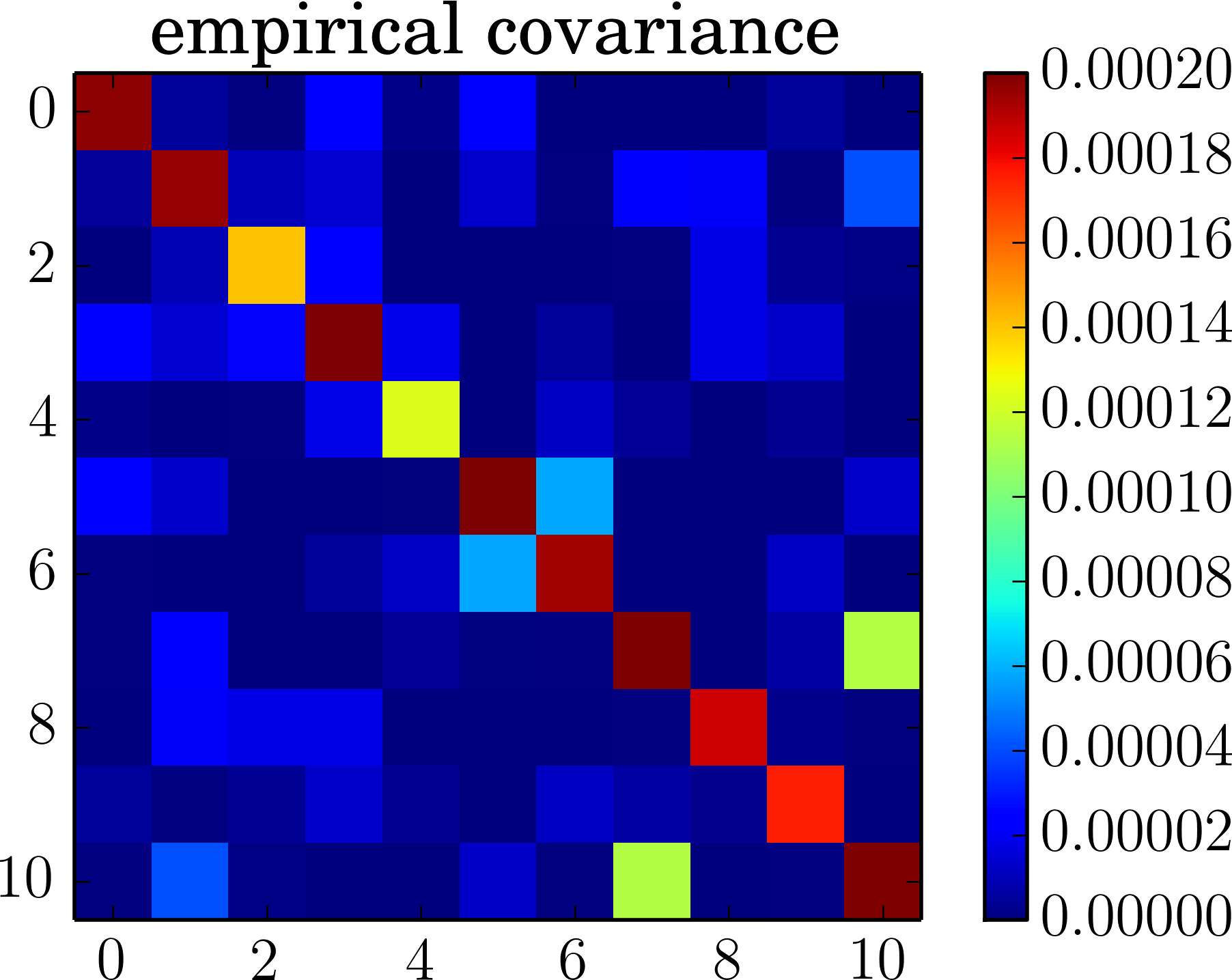}
\includegraphics[width=0.48\linewidth]{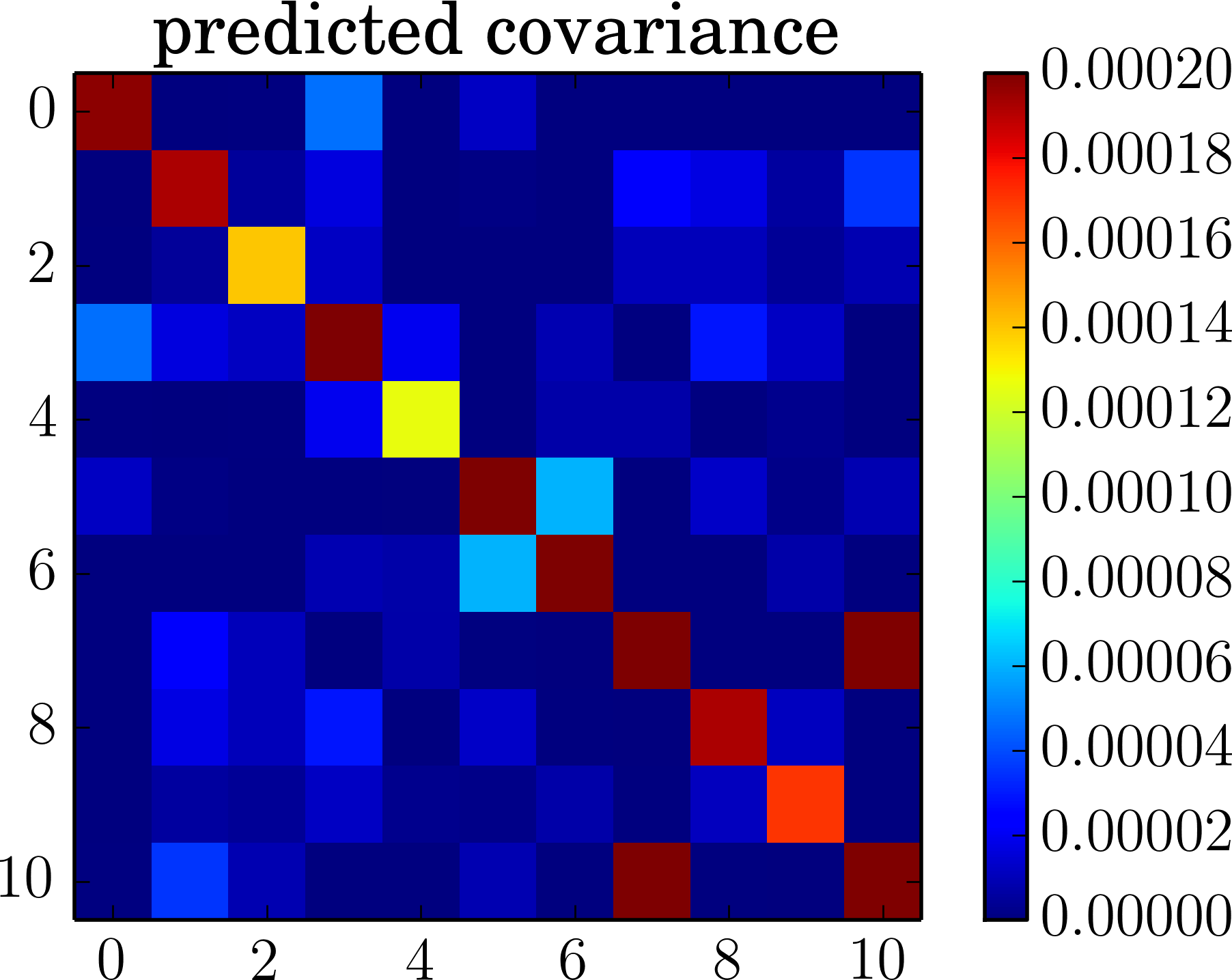}
\caption{Empirical and predicted covariances of the iterates of stochastic  gradient descent, where the prediction
is based on Eq.~\ref{eq:stationaryvariance} . 
We used linear regression on the wine quality data set as detailed in Section~\ref{sec:toy_data}. }
\label{fig:covariance}
\end{center}
\end{figure}
\vskip -0.2in

\subsection{Confirming the stationary distribution's covariance}
\label{sec:toy_data}
In this section, we confirm empirically that the stationary
distributions of SGD with KL-optimal constant learning rates are as predicted by the Ornstein-Uhlenbeck process.

\parhead{Data.} We first considered the following data sets.
(1) The \emph{Wine Quality Data Set}\footnote{P. Cortez, A. Cerdeira, F. Almeida, T. Matos and J. Reis, 'Wine Quality Data Set', UCI Machine Learning Repository.}, 
containing $N=4898$ instances, $11$ features, and one integer output variable (the wine rating). 
(2) A data set of \emph{Protein Tertiary Structure}\footnote{Prashant Singh Rana, 'Protein Tertiary Structure Data Set', UCI Machine Learning Repository.}, 
containing $N=45730$ instances, $8$ features and one output variable. 
(3) The \emph{Skin Segmentation Data Set}\footnote{Rajen Bhatt, Abhinav Dhall, 'Skin Segmentation Dataset', UCI Machine Learning Repository.}, 
containing $N=245057$ instances, $3$ features, and one binary output variable. We applied 
linear regression on data sets $1$ and $2$ and applied
logistic regression on data set $3$.
We rescaled the feature to unit length and used a mini-batch of sizes $S=100$, $S=100$  and $S=10000$, respectively. 
The quadratic regularizer was $1$. 
The constant learning rate was adjusted 
according to Eq.~\ref{eq:version_1}.

Fig.~\ref{fig:regression} shows two-dimensional projections of samples from the posterior (blue) and the stationary distribution (cyan), where
the directions were chosen two be the smallest and largest principal component of the posterior.
Both distributions are approximately Gaussian and centered around the maximum of the posterior.
To check our theoretical assumptions, we compared the covariance of the sampling distribution (yellow) against its predicted value based on the Ornstein-Uhlenbeck process (red),
where very good agreement was found. Since the predicted covariance is based on approximating SGD as a multivariate Ornstein-Uhlenbeck process, we conclude that our modeling assumptions are satisfied
to a very good extent. The unprojected 11-dimensional covariances on wine data are also compared in Fig.~\ref{fig:covariance}. 
The bottom row of  Fig.~\ref{fig:regression} shows the sampling distributions of black box variational inference (BBVI) using the reparametrization trick~\citep{kucukelbir2015automatic}. Our results show
that the approximation to the posterior given by constant SGD is not worse than the approximation given by BBVI. 

We also computed KL divergences between the posterior and stationary distributions of various algorithms:
constant SGD with KL-optimal learning rates and preconditioners, Stochastic Gradient Langevin Dynamics, Stochastic Gradient Fisher Scoring
(with and without diagonal approximation) and BBVI. 
For SG Fisher Scoring, we set the learning rate to $\epsilon^*$ of Eq.~\ref{eq:version_1}, while for Langevin dynamics
we chose the largest rate that yielded stable results ($\epsilon = \{10^{-3}, 10^{-6}, 10^{-5}\}$ for data sets $1$, $2$ and $3$, respectively). 
We found that constant SGD can compete in approximating the posterior with the MCMC algorithms under consideration.
This suggests that the most important factor is not the artificial noise involved in scalable MCMC, but rather the approximation of the
preconditioning matrix.

\subsection{Optimizing hyperparameters}
\label{sec:hyper_exp}

To test the hypothesis of Section~\ref{sec:hyperparameter_theory}, namely that constant
SGD as a variational algorithm allows for gradient-based hyperparameter learning,
we experimented with a
Bayesian multinomial logistic (a.k.a. softmax) regression model with
normal priors. The negative log-joint being optimized is
\begin{equation}
\begin{split}
\label{eq:logisticregression}
\mathcal{L}&\textstyle\equiv -\log p(y,\theta|x) = \frac{\lambda}{2}\sum_{d,k}\theta_{dk}^2 
- \frac{DK}{2}\log(\lambda) + \frac{DK}{2}\log2\pi
\\ &\textstyle + \sum_n \log\sum_k \exp\{\sum_d x_{nd}\theta_{dk}\}
- \sum_d x_{nd}\theta_{dy_n},
\end{split}
\end{equation}
where $n\in\{1,\ldots,N\}$ indexes examples, $d\in\{1,\ldots,D\}$
indexes features and $k\in\{1,\ldots,K\}$ indexes
classes. $x_n\in\mathbb{R}^D$ is the feature vector for the $n$th
example and $y_n\in\{1,\ldots,K\}$ is the class for that example.
Equation \ref{eq:logisticregression} has the degenerate maximizer
$\lambda=\infty$, $\theta=0$, which has infinite posterior density which
we hope to avoid in our approach.

\parhead{Data.} In all experiments, we applied this model to the MNIST dataset
(60000 training examples, 10000 test examples, 784
features) and the cover type dataset 
(500000
training examples, 81012 testing examples, 54 features).

\begin{figure}
\begin{center}
\includegraphics[width=0.9\linewidth]{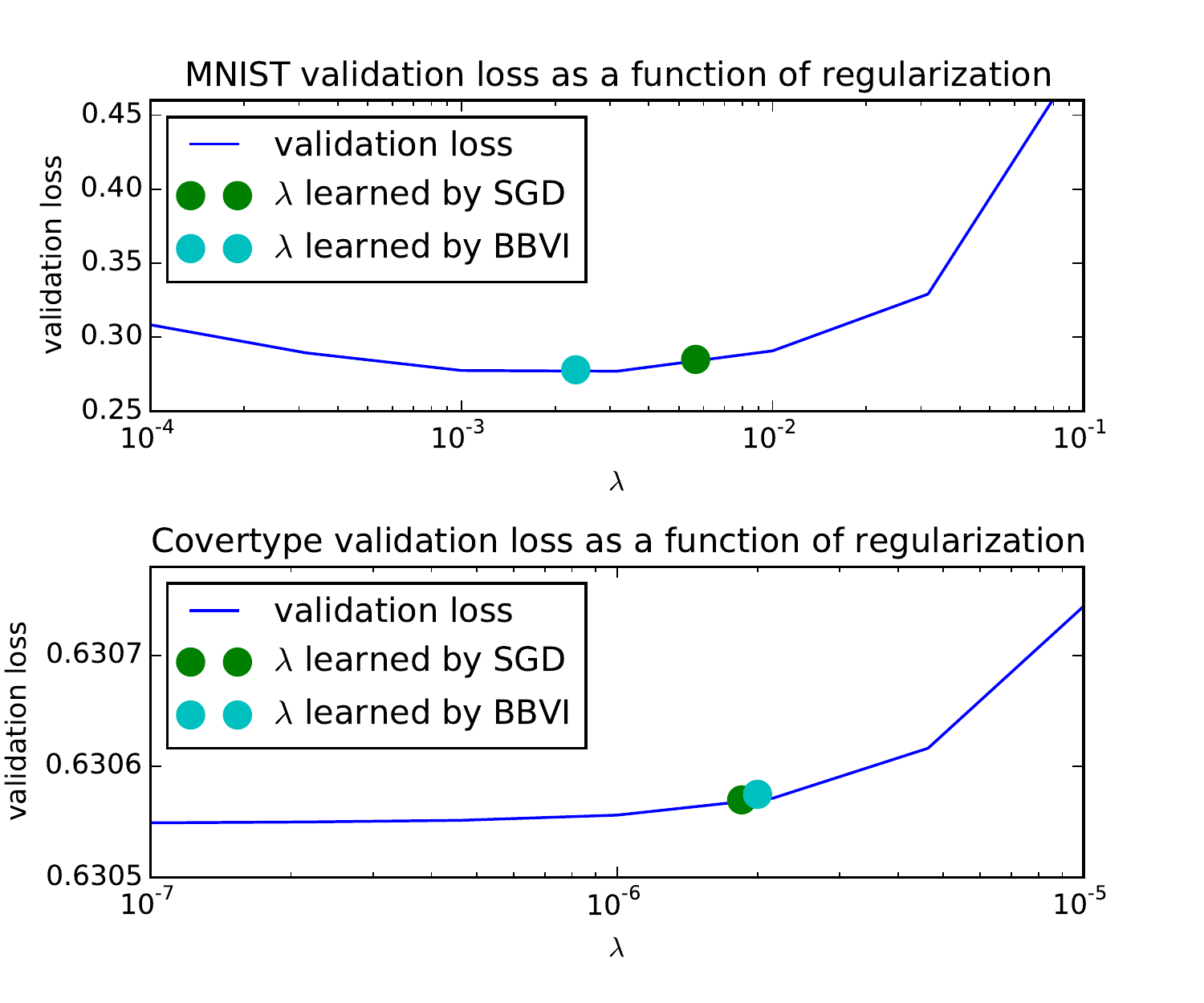}
\vskip-0.2in
\caption{Validation loss as a function of L2 regularization parameter
$\lambda$. Circles show the values of $\lambda$ that were automatically
selected by SGD and BBVI.}
\label{fig:optimizedlambda}
\end{center}
\end{figure}

Figure \ref{fig:optimizedlambda} shows the validation loss achieved by
maximizing equation \ref{eq:logisticregression} over $\theta$ for
various values of $\lambda$, as well as the values of $\lambda$
selected by SGD and BBVI. The results suggest that this approach can
be used as an inexpensive alternative to cross-validation or other
VEM methods for hyperparameter selection.

\input{related}

\section{Conclusions}
We analyzed new optimization goals of stochastic gradient descent
in the context of statistical machine learning. Instead of decreasing the learning rate to zero,
we ask for optimal constant learning rates such that Kullback-Leibler divergence between the stationary
distribution of SGD and the posterior is minimized. This goal leads to criteria for optimal learning rates, minibatch sizes
and preconditioning matrices. 
To be able to compute stationary distributions and KL divergences, we approximated
SGD in terms of a continuous-time stochastic process, leading to the Ornstein-Uhlenbeck process. We also presented
a novel analysis of stochastic gradient Fisher scoring.
Finally, we demonstrated that a simple SGD algorithm can
compete with stochastic variational methods at empirical Bayesian
hyperparameter optimization.

\bibliographystyle{apa} \bibliography{references}

\clearpage

\appendix

\section{Stationary Covariance}

The Ornstein-Uhlenbeck process has an analytic solution in terms of the stochastic integral (Gardiner et al., 1985), 
 \be
\label{eq:OU-solu}
\theta(t) = \exp(-At) \theta(0) + \sqrt{\textstyle\frac{\epsilon}{S}}\int_0^t \exp[-A(t-t')] B dW(t')
 \ee

Following Gardiner's book we derive an algebraic relation
for the stationary covariance of the multivariate Ornstein-Uhlenbeck process.
Define $\Sigma = \E[\theta(t)\theta(t)^\top]$. Using the formal solution for $\theta(t)$ given in the main paper, we find
\begin{align}
A \Sigma + \Sigma A^\top  &={\textstyle \frac{\epsilon}{S} } \int_{-\infty}^t A \exp[-A(t-t')] BB^\top  \exp[-A^\top(t-t')] dt' \n
					 &+{\textstyle \frac{\epsilon}{S} } \int_{-\infty}^t \exp[-A(t-t')] BB^\top  \exp[-A^\top(t-t')] dt' A^\top \n
					&= {\textstyle \frac{\epsilon}{S} } \int_{-\infty}^t  \frac{d}{dt'} \left(  \exp[-A(t-t')] BB^\top  \exp[-A^\top(t-t')]  \right) \n
					&={\textstyle \frac{\epsilon}{S} }  BB^\top. \nonumber
\end{align}
We used that the lower limit of the integral vanishes by the  positivity of the eigenvalues of $A$.

\section{Stochastic Gradient Fisher Scoring}
We start from the Ornstein-Uhlenbeck process
\be
d \Theta(t) & = & -HA \theta(t) dt + H \left[ B_{\epsilon/S} + E\right] dW(t) \n
		& = &  -A' \theta(t) dt +B' dW(t).
\ee
We defined $A'\equiv HA$ and $B'\equiv H \left[ B_{\epsilon/S} + E\right]$.
As derived in the paper, the variational bound is (up to a constant)
\be
KL \stackrel{\mathrm{c}}{=} \frac{N}{2}{\rm Tr}(A \Sigma) - \log (|NA|).
\ee
To evaluate it, the task is to remove the unknown covariance $\Sigma$ from the bound. To this end, as before, we use the identity 
for the stationary covariance $A'\Sigma + \Sigma A'^\top = B'B'^\top$.

The criterion for the stationary covariance is equivalent to
\be
HA\Sigma + \Sigma A H &=& \epsilon H BB^\top H + HEE^\top H^\top \n
\Leftrightarrow A\Sigma + H^{-1} \Sigma A H & = &  \epsilon BB^\top H + EE^\top H \n
\Rightarrow {\rm Tr}(A\Sigma) & = & \frac{1}{2}{\rm Tr}(H( \epsilon BB^\top + EE^\top))
\ee
We can re-parametrize the covariance as $\Sigma = TH$, such that $T$ is now unknown.
The KL divergence is therefore
\be
KL & = & -\frac{N}{2}{\rm Tr} (A\Sigma) + \log(|NA|) \n
       & = & \frac{N}{4}{\rm Tr}  (H( \epsilon BB^\top + EE^\top)) + \frac{1}{2} \log |T| \n
	& & + \frac{1}{2} \log |H| + \frac{1}{2} \log |NA| + \frac{D}{2},
\ee
which is the result we give in the main paper. 

\section{Square root preconditioning}

Finally, we analyze the case where we precondition with a matrix that
is proportional to the square root of the diagonal entries of the
noise covariance.

We define
\be
G &=& \sqrt{{\rm diag}(BB^\top)}
\ee
 as the diagonal matrix that contains square roots of the diagonal elements of the noise covariance.
We use an additional scalar learning rate $\epsilon$ .

\paragraph{Theorem (taking square roots).} 
\emph{Consider SGD preconditioned with $G^{-1}$ as defined above. 
Under the previous assumptions,  the constant learning rate which minimizes
KL divergence between the stationary distribution of this process and the posterior is}
\be
\label{eq:version_3}
\epsilon^* & = &\textstyle  \frac{2DS}{N{\rm Tr}(BB^\top G^{-1})}.
\ee
For the proof, we read off the appropriate KL divergence from the proof of Theorem 2 with $G^{-1}\equiv H$:
\begin{equation}
KL(q  || f )   \stackrel{\mathrm{c}}{=}   \textstyle\frac{\epsilon N}{2S}{\rm Tr}(BB^\top G^{-1}) - {\rm Tr}\log(G) +\textstyle\frac{D}{2} \log \textstyle\frac{\epsilon}{S} - \frac{1}{2}\log |\Sigma|
\end{equation}
Minimizing this KL divergence over the learning rate $\epsilon$ yields Eq.~\ref{eq:version_3} \qedsymbol.

\end{document}

%% file: introduction.tex
\section{Introduction}

Stochastic gradient descent (SGD) has become crucial to modern machine
learning. SGD optimizes a function by following noisy gradients with a
decreasing step size. The classical result
of~\citet{robbins1951stochastic} is that this procedure provably
reaches the optimum of the function (or local optimum, when it is
nonconvex). Recent studies investigate the merits of adaptive step
sizes~\citep{duchi2011adaptive, tieleman2012lecture}, gradient or 
iterate averaging~\cite{toulistowards,defossez2015averaged}, and constant
step-sizes~\cite{bach2013non, flammarion2015averaging}. Stochastic 
gradient descent has enabled efficient optimization with massive data.

Recently, stochastic gradients (SG) have also been used in the service of
scalable Bayesian Markov Chain Monte-Carlo (MCMC) methods, where the goal is
to generate samples from a
conditional distribution of latent variables given a data set.  In
Bayesian inference, we
assume a probabilistic model $p(\btheta, \bx)$ with data $\bx$ and
hidden variables $\btheta$; our goal is to approximate the posterior
\begin{align}
  \label{eq:intro-posterior}
  p(\btheta \g \bx) = \exp\{\log p(\btheta, \bx) - \log p(\bx)\}.
\end{align}
New scalable MCMC algorithms---such as SG Langevin
dynamics~\citep{welling2011bayesian}, SG Hamiltonian
Monte-Carlo~\citep{chen2014stochastic}, and SG Fisher
scoring~\citep{ahn2012bayesian}---employ stochastic gradients of
$\log p(\btheta, \bx)$ to improve convergence and computation of
existing sampling algorithms.  Also see~\citet{ma2015complete} for a
complete classification of these algorithms.

These methods all take precautions to sample from an asymptotically
exact posterior.  In contrast to this and specifically in the limit of
large data, we will show how to effectively use the simplest
stochastic gradient descent algorithm as a sensible \emph{approximate}
Bayesian inference method.  Specifically, we consider SGD with a
constant learning rate (constant SGD). Constant SGD first marches toward
an optimum of the objective function and then bounces around its vicinity 
because of the sampling noise in the gradient. (In contrast,
traditional SGD converges to the optimum by decreasing the
learning rate.)  Our analysis below rests on the idea that constant
SGD can be interpreted as a stochastic process with a stationary
distribution, one that is centered on the optimum and that has a
certain covariance structure.  The main idea is that we can use this
stationary distribution to approximate a posterior.

Here is how it works.  The particular profile of the stationary
distribution depends on the parameters of the algorithm---the constant
learning rate, the preconditioning matrix, and the minibatch size, all
of which affect the noise and the gradients.  Thus we can set
$\log p(\btheta, \bx)$ as the objective function and set the
parameters of constant SGD such that its stationary distribution is
close to the exact posterior
(Eq.~\ref{eq:intro-posterior}). Specifically, in the spirit of
variational Bayes~\citep{jordan1999introduction}, we set those parameters to
minimize the Kullback-Leibler (KL) divergence.  With those settings,
we can perform approximate  inference by simply running
constant SGD.

In more detail, we make the following contributions:
\begin{itemize}[leftmargin=*]

\item First, we develop a variational Bayesian view of stochastic
  gradient descent. Based on its interpretation as a
  continuous-time stochastic process---specifically a multivariate
  Ornstein-Uhlenbeck (OU) process~\citep{uhlenbeck1930theory,gardiner1985handbook}---we compute stationary
  distributions for a large class of SGD algorithms, all of which
  converge to a Gaussian distribution with a non-trivial covariance
  matrix. The stationary distribution is parameterized by the learning
  rate, minibatch size, and preconditioning matrix.

  Results about the multivariate OU process enable us to compute the
  KL divergence between the stationary distribution and the posterior analytically.
  Minimizing the KL,  
  we can relate the optimal step size or preconditioning matrix to the Hessian and noise covariances near the optimum.
  The resulting criteria  strongly resemble
  AdaGrad~\citep{duchi2011adaptive},
  RMSProp~\citep{tieleman2012lecture}, and classical Fisher
  scoring~\citep{longford1987fast}.  
  We demonstrate how these different
  optimization methods compare, when used for approximate inference.

\item Then, we analyze scalable MCMC
  algorithms. Specifically, we use the stochastic process perspective
  to compute the stationary sampling distribution of stochastic
  gradient Fisher scoring~\citep{ahn2012bayesian}.   
  The view from the multivariate OU process reveals a simple
  justification for this method: we show that the preconditioning
  matrix suggested in SGFS is indeed optimal. We also derive a
  criterion for the free noise parameter in SGFS such as to enhance numerical stability, and we show how the
  stationary distribution is modified when the preconditioner is
  approximated with a diagonal matrix (as is often done in practice
  for high-dimensional problems).

\item Finally, we show how using SGD with a constant learning rate confers
  an important practical advantage: it allows simultaneous inference of the
  posterior and optimization of meta-level parameters, such as
  hyperparameters in a Bayesian model.  We demonstrate this technique
  on a Bayesian multinomial logistic regression model with normal
  priors.
\end{itemize}

Our paper is organized as follows. In section~\ref{sec:formalism} we
review the continuous-time limit of SGD, showing that it can be
interpreted as an OU process. In section~\ref{sec:consequences} we
present consequences of this perspective: the interpretation of SGD as
variational Bayes  and results around
stochastic gradient Fisher Scoring~\cite{ahn2012bayesian}. In the
empirical study (section \ref{sec:experiments}), we show that our theoretical
assumptions are satisfied for different models, and that we can use
SGD to perform gradient-based hyperparameter optimization.

%% file: framework.tex
\section{Continuous-Time Limit Revisited}
\label{sec:formalism}

We first review the theoretical framework that we use throughout the
paper.  Our goal is to characterize the behavior of SGD when using a
constant step size.  To do this, we approximate SGD with a
continuous-time stochastic
process~\citep{kushner2003stochastic,ljung2012stochastic}.

\subsection{Problem setup}

Consider loss functions of the following form:
\begin{align}
\label{eq:loss-one}
{\cal L}(\theta)
= {\textstyle\frac{1}{N} \sum_{n=1}^N} \ell_n(\theta), \quad
g(\theta)\equiv \nabla_\theta \cal{L}(\theta).
\end{align}
Such loss functions are common in machine learning, where
${\cal L}(\theta) \equiv {\cal L}(\theta,x)$ is a loss function that
depends on data $x$ and parameters $\theta$. Each
$ \ell_n(\theta)\equiv \ell(\theta,x_n)$ is the contribution to the
overall loss from a single observation.  For example, when
finding a maximum-a-posteriori estimate of a model, the contributions to the
loss may be
\begin{align}
\label{eq:loss-two}
\ell_n(\theta) = - \log p(x_n \,|\,
\theta) + \frac{1}{N} \log p(\theta),
\end{align}
where $p(x_n \,|\, \theta)$ is the likelihood and $p(\theta)$ is the prior.
For simpler notation, we will suppress the dependence of the loss on
the data.

From this loss we construct stochastic gradients.  Let ${\cal S}$ be a
set of $S$ random indices drawn uniformly at random from the set
$\{1,\ldots, N\}$.  This set indexes functions $\ell_n(\theta)$, and
we call $\cal{S}$ a ``minibatch'' of size $S$.  Based on the
minibatch, we used the indexed functions to form a stochastic estimate
of the loss and a stochastic gradient, 
\begin{align}
\label{eq:SG}
\hat{\cal L}_S(\theta) = \textstyle\frac{1}{S} \sum_{n \in {\cal S}}\, \ell_n(\theta), \quad \hat{g}_S(\theta) = \nabla_\theta\hat{\cal L}_S(\theta).
\end{align}
In expectation the stochastic gradient is the full gradient, i.e., $g(\theta) = {\mathbb E}[\hat{g}_S(\theta)]$.
We use this stochastic gradient in the SGD update
\begin{align}
\label{eq:SGD}
\theta(t+1)  =  \theta(t) - \epsilon \, \hat{g}_S(\theta(t)).
\end{align}

Equations~\ref{eq:SG} and~\ref{eq:SGD} define the discrete-time
process that SGD simulates from. We will approximate it with a
continuous-time process that is easier to analyze.
 
\subsection{SGD as a Ornstein-Uhlenbeck process}

We now show how to approximate the discrete-time Eq.~\ref{eq:SGD} with
the continuous-time Ornstein-Uhlenbeck
process~\citep{uhlenbeck1930theory}.  This leads to the stochastic
  differential equation below in Eq.~\ref{eq:OU-process}.  To justify
  the approximation, we make four assumptions.  We verify its accuracy
  in Section~\ref{sec:experiments}.

\parhead{Assumption 1.} Observe that the stochastic gradient is a sum
of $S$ independent, uniformly sampled contributions.  Invoking the
central limit theorem, we assume that the gradient noise is Gaussian
with variance~$\propto 1/S$:
\begin{align}
\label{eq:SGD2}
\hat{g}_S(\theta)  \approx  g(\theta) + {\textstyle \frac{1}{\sqrt{S}}}\Delta g(\theta),\quad  \Delta g(\theta) \sim {\cal N}(0,C(\theta)). 
\end{align}

\parhead{Assumption 2.} We assume that the noise covariance is
approximately constant. Further, we decompose the constant noise
covariance into a product of two constant matrices: 
\begin{align}
\label{eq:CBB}
C  = B B^\top.
\end{align}
This assumption is justified when the iterates of SGD are confined to
a small enough region around a local optimum of the loss that the noise
covariance does not vary significantly in that region.

\parhead{Assumption 3.}  We now define
$\Delta \theta(t) = \theta (t+1) - \theta(t)$ and combine
Eqs.~\ref{eq:SGD},~\ref{eq:SGD2}, and~\ref{eq:CBB} to rewrite the
process as
\begin{align}
\Delta \theta(t)  =  -\epsilon \, g(\theta(t))
+\sqrt{\textstyle \frac{\epsilon}{S}}B \,\Delta W, \quad \Delta W \sim
{\cal N}\left(0,\epsilon{\bf I}\right).
\end{align}
This is a discretization of the following continuous-time stochastic differential equation:
\footnote{We performed the conventional substitution rules when discretizing a continuous-time stochastic
process. These substitution rules are $\Delta \theta(t) \rightarrow d\theta(t) $, $\epsilon \rightarrow dt$ and $\Delta W \rightarrow dW$, see e.g.~\citep{gardiner1985handbook}.}
\begin{align}
\label{eq:Langevin}
d \theta(t)  =  -g(\theta) dt +\sqrt{\textstyle\frac{\epsilon}{S}}B  \,  dW(t).
\end{align}
We assume that this continuous-time limit is approximately justified and that we can neglect the discretization errors.

\parhead{Assumption 4.} Finally, we assume that the stationary
distribution of the iterates is constrained to a region where the loss
is well approximated by a quadratic function,
\begin{align}
{\cal L}(\theta)  =  \textstyle\frac{1}{2} \, \theta^\top A \theta.
\end{align}
(Without loss of generality, we assume that a minimum of the loss is at $\theta=0$.)
This assumption makes sense when the loss function is smooth and
the stochastic process
reaches a low-variance quasi-stationary distribution around a deep local minimum.
The exit time of a stochastic process is typically
exponential in the height of the barriers between minima, which can make local optima very stable even in the presence of noise~\citep{kramers1940brownian}.

\begin{figure}[t!]
\begin{center}
\includegraphics[width=.49\linewidth]{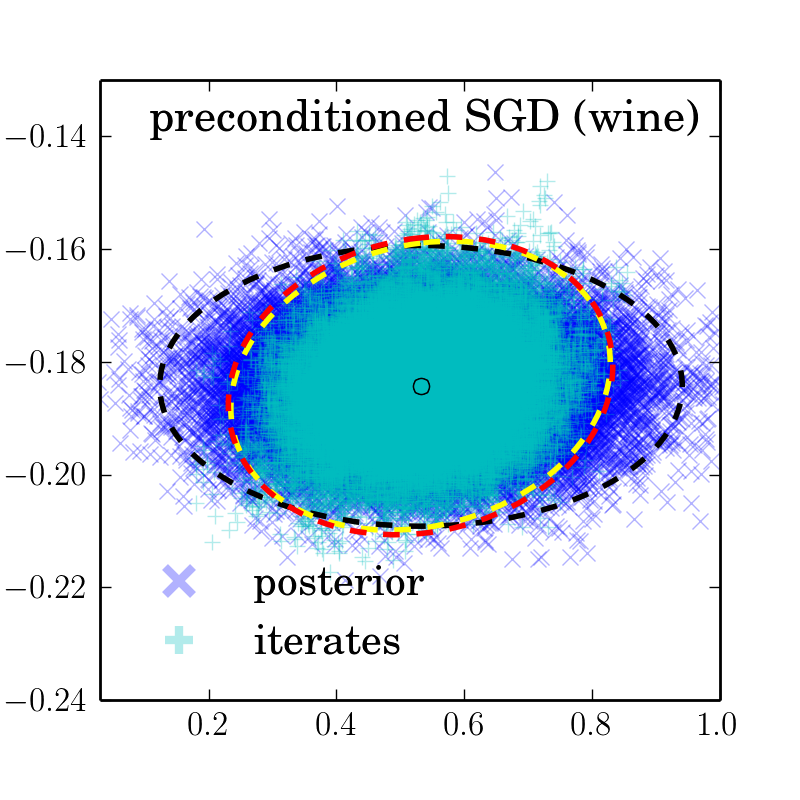} 
\includegraphics[width=.49\linewidth]{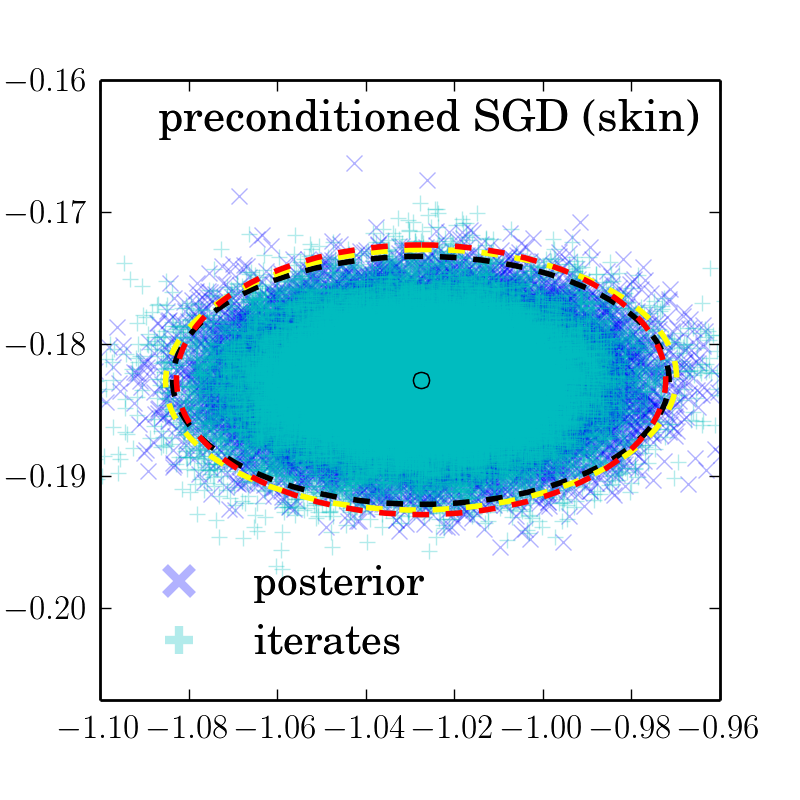} 
\includegraphics[width=.49\linewidth]{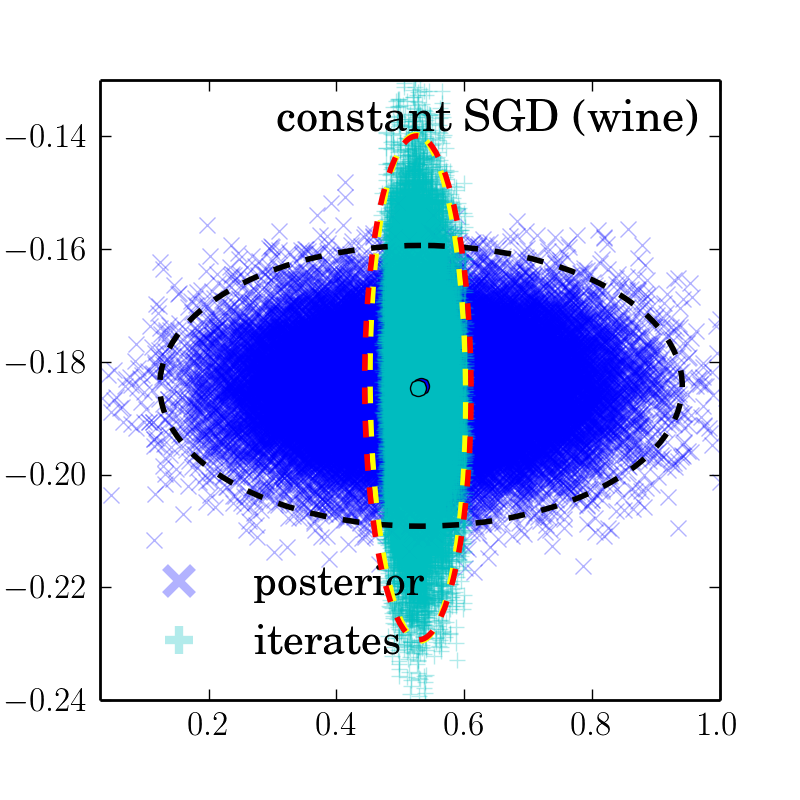} 
\includegraphics[width=.49\linewidth]{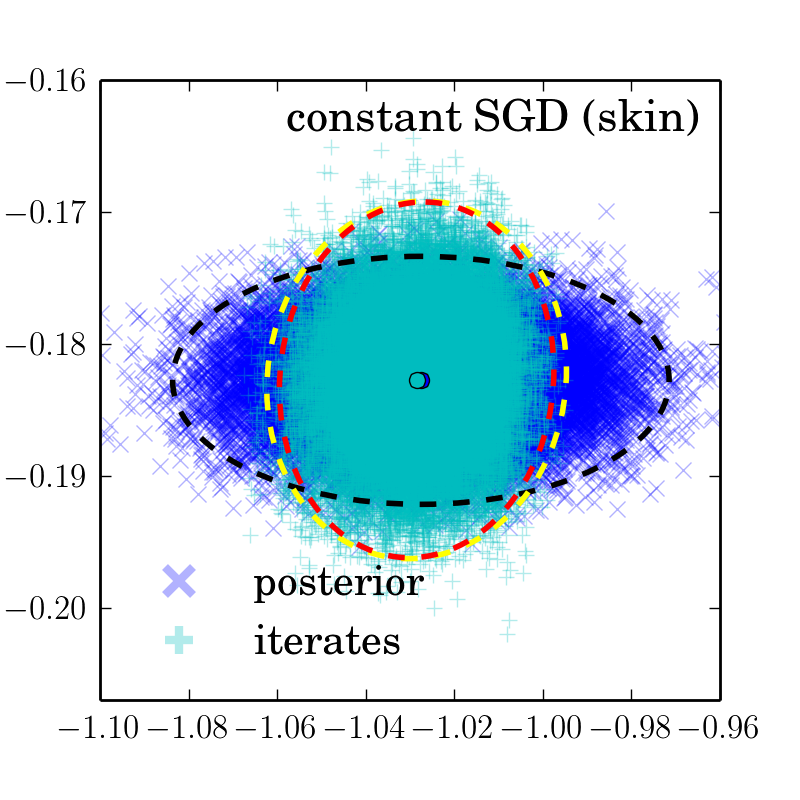} 
\includegraphics[width=.49\linewidth]{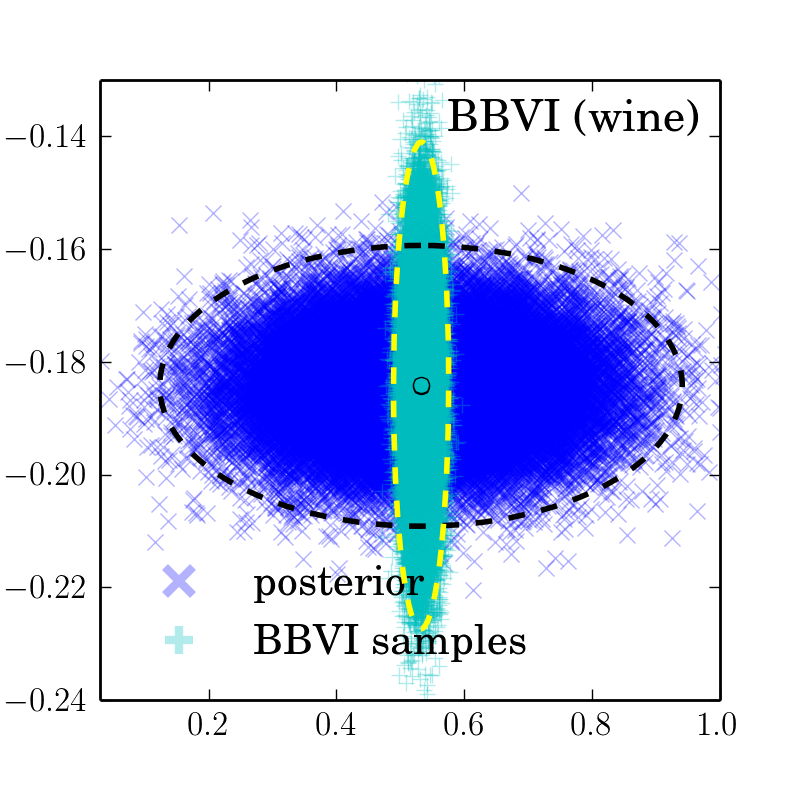} 
\includegraphics[width=.49\linewidth]{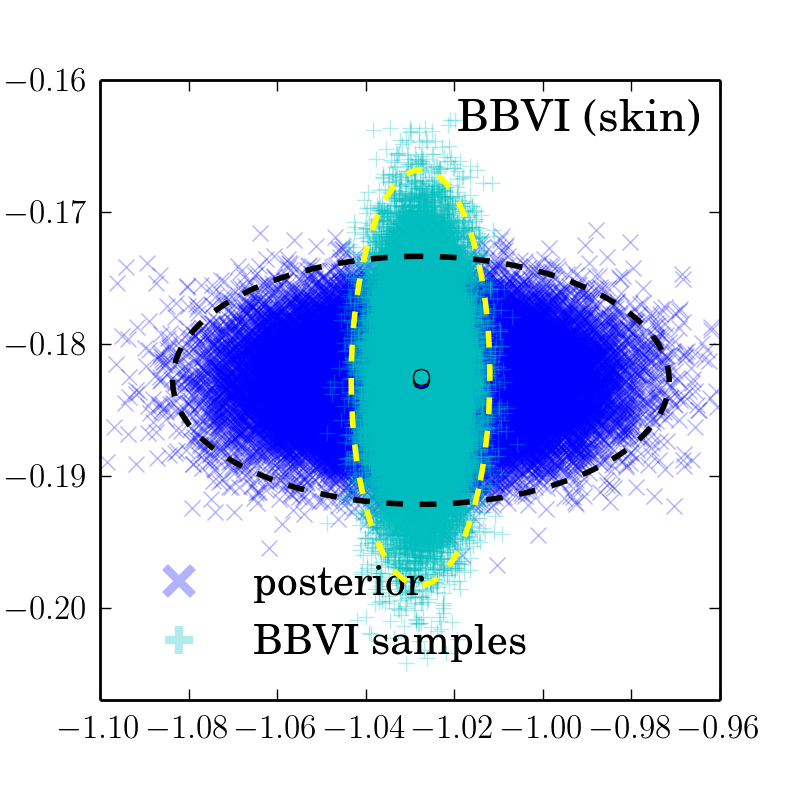} 
\caption{
	Posterior distribution $f(\theta) \propto \exp\left\{-N{\cal L}(\theta)\right\}$ (blue)
	and stationary sampling distributions $q(\theta)$ of the iterates of SGD (cyan) or black box variational inference (BBVI). Columns: 
	linear regression (left) and
	logistic regression (right) discussed in Section~\ref{sec:experiments}. Rows: 
	full-rank preconditioned constant SGD (top), constant SGD (middle), and BBVI{\small~\citep{kucukelbir2015automatic}} (bottom). We show projections
	on the smallest and largest principal component of the posterior. The plot also shows
	the empirical covariances (3 standard deviations) of the posterior (black), the covariance of
	the samples (yellow), and their prediction (red) in
	terms of the Ornstein-Uhlenbeck process, Eq.~\ref{eq:stationaryvariance}.
	}
\label{fig:regression}
\end{center}
\end{figure}

\parhead{SGD as an Ornstein-Uhlenbeck process.}  For what follows,
define $B_{\epsilon/S} = \sqrt{\textstyle\frac{\epsilon}{S}}B$.  The
four assumptions above result in a specific kind of stochastic process, the
multivariate \emph{Ornstein-Uhlenbeck}
process~\citep{uhlenbeck1930theory}.  It is
\begin{align}
\label{eq:OU-process}
d \theta(t)  =  -A\, \theta(t) dt + B_{\epsilon/S}\, dW(t) 
\end{align}

This connection helps us analyze properties of SGD because the
Ornstein-Uhlenbeck process has an analytic stationary distribution
$q(\theta)$ that is Gaussian. This distribution will be the core analytic tool
of this paper: 
\begin{align}
\label{eq:stationarydistribution}
q(\theta)  \propto 
\exp\left\{-\textstyle\frac{1}{2}\theta^\top \Sigma^{-1}\theta\right\}. 
\end{align}
The covariance $\Sigma$ satisfies
\begin{align}
\label{eq:stationaryvariance}
\Sigma A^\top + A \Sigma = \textstyle\frac{\epsilon}{S}BB^\top.
\end{align}

Without explicitly solving this equation, we see that the resulting
covariance is proportional to the learning rate $\epsilon$ and
inversely proportional to
the magnitude of $A$
and minibatch size
$S$.  (More details are in the Appendix.)  This characterizes the
stationary distribution of running SGD with a constant step size.

%% file: inference.tex
\begin{table}
\begin{center}
\begin{tabular}{|l||c|c|c|}
\hline
Method                   & Wine    & Skin  & Protein\\ \hline\hline
constant SGD       & 18.7                & 0.471   & 1000.9 \\ \hline
constant SGD-d   & 14.0                & 0.921   & 678.4 \\ \hline
constant SGD-f    &  0.7                 & 0.005   & 1.8 \\ \hline
SGLD~{\small \citep{welling2011bayesian}  }                  &  2.9                 &0.905   & 4.5 \\ \hline
SGFS-d~{\small\citep{ahn2012bayesian}} &  12.8                 &0.864  & 597.4  \\ \hline
SGFS-f~{\small\citep{ahn2012bayesian}}    &  0.8                 &0.005   & 1.3  \\ \hline
BBVI{\small~\citep{kucukelbir2015automatic}}  &  44.7                & 5.74       & 478.1\\ \hline
\end{tabular}
\end{center}
\caption{ KL divergences between the posterior and stationary sampling distributions applied to the data sets discussed in Section~\ref{sec:toy_data}.
We compared constant SGD without preconditioning and with diagonal (-d) and full rank (-f) preconditioning
against Stochastic Gradient Langevin Dynamics and Stochastic Gradient Fisher Scoring (SGFS) with diagonal (-d) and full rank (-f) preconditioning, and BBVI.
}
\end{table}

\section{SGD as Approximate Inference}
\label{sec:consequences}

We discussed a continuous-time interpretation of SGD with a constant
step size (constant SGD).  We now discuss how to use constant SGD as
an approximate inference algorithm.  To repeat the set-up from the
introduction, consider a probabilistic model $p(\btheta, \bx)$ with
data $\bx$ and hidden variables $\btheta$; our goal is to approximate
the posterior in Eq.~\ref{eq:intro-posterior}.

We set the loss to be proportional to the negative log-joint distribution
(Eqs.~\ref{eq:loss-one} and \ref{eq:loss-two}), which equals the
posterior up to an additive constant.  The classical goal of SGD is to
minimize this loss, leading us to a maximum-a-posteriori point estimate of the
parameters. This is how SGD is used in many statistical models,
including logistic regression, linear regression, matrix
factorization, neural network classifiers, and regressors.  In
contrast, our goal here is to tune the parameters of SGD such that we
approximate the posterior with its stationary distribution. Thus we
use SGD as a posterior inference algorithm.

Fig.~\ref{fig:regression} shows an example. Here we illustrate two
Bayesian posteriors---from a linear regression problem (left) and a logistic
regression problem (right)---along with iterates from a constant SGD
algorithm. In these figures, we set the parameters of the optimization
to values that minimize the Kullback-Leibler (KL) divergence between
the stationary distribution of the OU process and the
posterior---these results come from our theorems below. The top plots
optimize both a preconditioning matrix and the step size; the middle
plots optimize only the step size. (The middle plots are from a more
efficient algorithm, but it is less accurate.) We can see that the
stationary distribution of constant SGD can be made close to the exact
posterior.

Fig.~\ref{fig:regression} also compares the empirical covariance of the
iterates with the predicted covariance in terms of
Eq.~\ref{eq:stationaryvariance}. The close match supports the
assumptions of Sec.~\ref{sec:formalism}.

We will use this perspective in two ways.  First, we develop optimal
algorithmic conditions for constant SGD to best approximate the
posterior, connecting to well-known results around adaptive learning
rates and preconditioners.  Second, we use it to analyze Stochastic
Gradient Fisher Scoring~\citep{ahn2012bayesian}, both in its exact
form and its more efficient approximate form.

\subsection{Constant stochastic gradient descent}
\label{sec:SGD_bare}

First, we show how to tune constant SGD's parameters to minimize the
KL divergence to the posterior; this is a type of variational
inference~\citep{Jordan:1999}. 
Based on this analysis, we
derive three versions of constant SGD---one with a constant step size,
one with a full preconditioning matrix, and one with a diagonal
preconditioning matrix.  Each one yields samples from an approximate
posterior, and each trades of efficiency and accuracy in a different
way.  Finally, we show how to use these 
algorithms to learn hyperparameters.

Assumption 4 from Sec.~\ref{sec:formalism} says that the posterior is
approximately Gaussian in the region that the stationary distribution
focuses on,
\begin{align}
\label{eq:posterior}
f(\theta)  \propto   \exp\left\{-\textstyle\frac{N}{2}\theta^\top A \theta \right\}.
\end{align}
(The scalar $N$ corrects the averaging in equation \ref{eq:loss-one}.)
In setting the parameters of SGD, we minimize the KL divergence
between the posterior $f(\theta)$ and the stationary distribution
$q(\theta)$ (Eqs.~\ref{eq:stationarydistribution} and
\ref{eq:stationaryvariance}) as a function of the learning rate
$\epsilon$ and minibatch size $S$.  We can optionally include a
\emph{preconditioning matrix} $H$, i.e. a matrix that premultiplies
the stochastic gradient to modify its convergence behavior.

Hence, we minimize
\begin{align}
\{\epsilon^{*},S^{*},H^{*}\}  = 
\arg\min_{\epsilon,S,H} KL(q(\theta) \mid\mid f(\theta)).
\end{align}
The distributions $f(\theta)$ and $q(\theta)$ are both
Gaussians. Their means coincide, at the minimum of the loss, and so
their KL divergence is
\begin{align}
\label{eq:KL1}
KL(q  || f ) & =  \E_{q(\theta)}[\log f(\theta)] -  \E_{q(\theta)}[\log q(\theta)]  \\
& =  \textstyle\frac{1}{2}\left( N {\rm Tr}(A\Sigma) - \log |NA| -
  \log |\Sigma| - D \right),\nonumber
\end{align} where $|\cdot|$ is the
determinant and $D$ is the dimension of $\theta$.

We suggest three variants of constant SGD that generate samples from
an approximate posterior.

\paragraph{Theorem 1 (constant SGD).} \emph{Under assumptions
  [A1-A4], the constant learning rate which minimizes KL divergence
  from the stationary distribution of SGD to the posterior is}
\begin{align}
\label{eq:version_1}
\epsilon^*  =   \textstyle\frac{2 D S }{N {\rm Tr}(BB^\top)}.
\end{align}
To prove this claim, we face the problem that the covariance of the
stationary distribution depends indirectly on $\epsilon$ through
Eq.~\ref{eq:stationaryvariance}. Inspecting this equation reveals
that $\Sigma_0 \equiv \textstyle\frac{S}{\epsilon}\Sigma$ is
independent of $S$ and $\epsilon$. This simplifies the entropy term
$\log |\Sigma| = D \log (\epsilon/S) + \log |\Sigma_0|$. Since
$\Sigma_0$ is constant, we can neglect it when minimizing KL
divergence.

We also need to simplify the term ${\rm Tr}(A\Sigma)$, which still
depends on $\epsilon$ and $S$ through $\Sigma$. To do this, we again use
Eq.~\ref{eq:stationaryvariance}, from which follows that
${\rm Tr}(A\Sigma) = \frac{1}{2}({\rm Tr}(A\Sigma)+{\rm Tr}(\Sigma
A^\top)) = \frac{\epsilon}{2S}{\rm Tr}(BB^\top)$.
The KL divergence is therefore, up to constant terms,
\begin{align}
KL(q  || f ) \stackrel{\mathrm{c}}{=} \textstyle\frac{\epsilon\, N}{2S} {\rm Tr}(BB^\top) - D \log (\epsilon/S)
\end{align}
Minimizing KL divergence over $\epsilon/S$ results in Eq.~\ref{eq:version_1} for the optimal learning rate. \qedsymbol

Theorem 1 suggests that the learning rate should be chosen inversely
proportional to the average of diagonal entries of the noise covariance.
We can also precondition SGD with a diagonal matrix $H$.  This gives more tuning
parameters to better approximate the posterior.  Under the same
assumptions, we ask for the optimal diagonal preconditioner.

\paragraph{Theorem 2 (preconditioned constant SGD).}
\emph{The constant preconditioning matrix for SGD that
  minimizes KL divergence from the sampling distribution to the
  posterior is} \begin{align}
\label{eq:version_2}
H^* = \textstyle \frac{ 2S}{ \epsilon N} (BB^\top)^{-1}
\end{align}
To prove this claim, we need the Ornstein-Uhlenbeck process which corresponds
to preconditioned SGD. Preconditioning Eq.~\ref{eq:OU-process} with H results in
\begin{align}
d \theta(t)  =  - HA \,\theta(t) dt + H B_{\epsilon/S} dW(t).
\end{align}
All our results carry over after substituting $A  \leftarrow  HA, \;  B  \leftarrow  HB$.
Eq.~\ref{eq:stationaryvariance}, after the transformation and
multiplication by $H^{-1}$ from the left, becomes
\begin{align}
A \Sigma   + H^{-1} \Sigma A^\top H= \textstyle\frac{\epsilon}{S} BB^\top H
\end{align}
Using the cyclic property of the trace, this implies that ${\rm Tr}(A\Sigma) = \textstyle\frac{1}{2}({\rm Tr}(A\Sigma) + {\rm Tr}(H^{-1}A\Sigma H) = \textstyle\frac{\epsilon}{2S}{\rm Tr}(BB^\top H)$.
Hence up to constant terms, the  KL divergence is
\begin{align}
\label{eq:mod_KL}
KL(q  || f ) & \stackrel{\mathrm{c}}{=}   \textstyle\frac{\epsilon N}{2S}{\rm Tr}(BB^\top H) + \textstyle\frac{1}{2} \log \left(\textstyle\frac{\epsilon}{S}|H \Sigma^{-1}H| \right) \\
		 & =   \textstyle\frac{\epsilon N}{2S}{\rm Tr}(BB^\top H) + {\rm Tr}\log(H) +\textstyle\frac{D}{2} \log \textstyle\frac{\epsilon}{S} -\textstyle\frac{1}{2}  \log |\Sigma|. \nonumber
\end{align}
(We used that $\log (\det H) = {\rm Tr} \log H$.)
Taking derivatives with respect to the entries of $H$ results in Eq.~\ref{eq:version_2}. \qedsymbol

In high-dimensional applications where working with large dense matrices is impractical, the preconditioner may be constrained to be diagonal.
The following corollary is a direct consequence of
Eq.~\ref{eq:mod_KL} when constraining the
preconditioner  to be diagonal:

\paragraph{Corollary 1}
\emph{The optimal diagonal preconditioning matrix for SGD that
  minimizes KL divergence is}
\begin{align}
H_{kk}^* = \textstyle \frac{ 2S}{ \epsilon N BB_{kk}^\top}.
\end{align}

We showed that the optimal diagonal preconditioner is the inverse of
the diagonal part of the noise matrix.  Similar preconditioning
matrices have been suggested earlier in optimal control theory based
on very different arguments, see~\citep{widrow1985adaptive}.  Our
result also relates to AdaGrad and its
relatives~\citep{duchi2011adaptive, tieleman2012lecture}, which also
adjust the preconditioner based on the square root of the diagonal entries of the noise covariance.
In the supplement
we derive an optimal global learning rate for AdaGrad-style diagonal preconditioners.

In Sec.~\ref{sec:experiments}, we compare three versions of
constant SGD for approximate posterior inference: one with a scalar
step size, one with a diagonal preconditioner, and one with a dense
preconditioner.

\subsection{Stochastic Gradient Fisher Scoring}
\label{sec:fisher}

We now investigate Stochastic Gradient Fisher
Scoring~\citep{ahn2012bayesian}, a scalable Bayesian
MCMC algorithm.  We use the variational perspective to rederive the Fisher
scoring update and identify it as optimal. We also analyze the
sampling distribution of the truncated algorithm, one with diagonal
preconditioning (as it is used in practice), and quantify the bias
that this induces.

The basic idea here is that the stochastic gradient is preconditioned
and additional noise is added to the updates such that the algorithm
approximately samples from the Bayesian posterior.  More precisely,
the update is \begin{align}
\label{eq:SGFS}
\theta(t+1)  =  \theta(t) - \epsilon H \, \hat{g}(\theta(t)) + \sqrt{\epsilon} H E \, W(t).
\end{align}
The matrix $H$ is a preconditioner and $E W(t)$ is Gaussian noise; we
control the preconditioner and the covariance $EE^\top$ of the
noise. Stochastic gradient Fisher scoring suggests a preconditioning
matrix $H$ that leads to samples from the posterior even if the
learning rate $\epsilon$ is not asymptotically small.  We show here
that this preconditioner follows from our variational analysis.

\paragraph{Theorem 3 (SGFS)}
\emph{Under assumptions A1-A4, the preconditioning matrix $H$  in Eq.~\ref{eq:SGFS} that
minimizes KL divergence between the stationary distribution of SGFS and the posterior is
}
\begin{align}
\label{eq:fisher_H}
\textstyle H^{*}  =  \frac{2}{N}( \epsilon BB^\top + EE^\top)^{-1}.
\end{align}
To prove the claim, we go through the steps of section~\ref{sec:formalism} to derive the corresponding Ornstein-Uhlenbeck process,
$d \theta(t)  =  -HA \theta(t) dt + H \left[  B_{\epsilon} + E\right] dW(t).$
For simplicity, we have set the minibatch size $S$ to $1$, hence $B_{\epsilon}\equiv \sqrt{\epsilon}B$.
In the appendix, we derive the following KL divergence between the posterior and the sampling distribution:
$
KL(q||p)  =  - \textstyle\frac{N}{4}{\rm Tr}  (H(B_{\epsilon}B^\top_{\epsilon} + EE^\top)) + \textstyle\frac{1}{2} \log |T| + \textstyle\frac{1}{2} \log |H| + \textstyle\frac{1}{2} \log |NA| + \textstyle\frac{D}{2}.
$
We can now minimize this KL divergence over the parameters $H$ and $E$.
When $E$ is given, minimizing over $H$ gives Eq.~\ref{eq:fisher_H}  \qedsymbol.

The solution given in Eq.~\ref{eq:fisher_H} not only minimizes the KL
divergence, but makes it $0$, meaning that the stationary
sampling distribution \emph{is} the posterior. This solution
corresponds to the suggested Fisher Scoring update in the idealized
case when the sampling noise distribution is estimated
perfectly~\citep{ahn2012bayesian}. Through this update, the algorithm
thus generates posterior samples without decreasing the
learning rate to zero.  (This is in contrast to Stochastic Gradient
Langevin Dynamics by~\citet{welling2011bayesian}.)

In practice, however, SGFS is often used with a diagonal approximation of
the preconditioning matrix~\citep{ahn2012bayesian,ma2015complete}.
However, researchers have not explored how the stationary distribution
is affected by this truncation, which makes the algorithm only
approximately Bayesian.  
We can quantify its deviation from the exact posterior and we derive the optimal 
diagonal preconditioner.

\paragraph{Corollary 2 (approximate SGFS).}
\emph{
When approximating the Fisher scoring preconditioning matrix by a diagonal matrix or a scalar, respectively,
then
$H_{kk}^{*}  =  \frac{2}{N}( \epsilon BB^\top_{kk} + EE^\top_{kk})^{-1}$ and $H^*_{scalar} = \frac{2D}{N} (\sum_k [ \epsilon BB^\top_{kk} + EE^\top_{kk}])^{-1}$, respectively.
}

This follows from the KL divergence in the proof of theorem 3.  

Note that we have not made any assumptions about the noise covariance $E$.
To keep the algorithm stable, it may make sense to impose a 
maximum step size $h^{max}$, so that $H_{kk}<h^{max}$. We can satisfy 
Eq.~\ref{eq:fisher_H} by choosing $H_{kk} = h^{max} = \frac{2}{N} ( \epsilon BB^\top_{kk} + EE^\top_{kk})^{-1}$.
Solving for $E$ yields
\begin{align}
E_{kk}  =  \textstyle\frac{2}{h^{max}N} -  \epsilon BB^\top_{kk}.
\end{align}
Hence, to keep the learning rates bounded in favor of
stability, one can inject noise in dimensions where the variance of
the gradient is too small.  This guideline is opposite to the
advice of choosing $B$ proportional to $E$, as suggested
by~\citet{ahn2012bayesian}, but follows naturally from the variational analysis.

\subsection{Implications for hyperparameter optimization}
\label{sec:hyperparameter_theory}

One of the major benefits to the Bayesian approach is the ability to
fit hyperparameters to data without expensive cross-validation runs by
placing hyperpriors on those hyperparameters. Empirical Bayesian
methods fit hyperparameters by finding the hyperparameters that
maximize the \emph{marginal} likelihood of the data, integrating out
the main model parameters:
\begin{equation}
\begin{split}
\textstyle \lambda^\star = \arg\max_{\lambda}\log
p(y|x,\lambda)=\arg\max_{\lambda}\log\int_\theta p(y,\theta|x,\lambda)d\theta.
\nonumber
\end{split}
\end{equation}
When this marginal log-likelihood is intractable, a common approach is
to use \emph{variational expectation-maximization (VEM)}
\citep{Bishop:2006}, which iteratively optimizes a variational lower bound on the
marginal log-likelihood over $\lambda$.  If we approximate the
posterior $p(\theta|x,y,\lambda)$ with some distribution $q(\theta)$,
then VEM tries to find a value for $\lambda$ that maximizes the
expected log-joint probability $\mathbb{E}_q[\log
  p(\theta,y|x,\lambda)]$.

If we interpret the stationary distribution of SGD as a variational
approximation to a model's posterior, then we can justify jointly
optimizing parameters and hyperparameters as a VEM algorithm.
This should avoid degenerate solutions, as long as we use
the learning rates and preconditioners derived above.
In the experimental section, we show that gradient-based hyperparameter
learning is a cheap alternative to cross-validation in constant SGD.

%% file: related.tex
\section{Related Work}
Our paper relates to Bayesian inference and stochastic optimization.

\parhead{Scalable MCMC.} Recent work in Bayesian statistics focuses on
making MCMC sampling algorithms scalable by using stochastic
gradients. In particular, \citet{welling2011bayesian} developed
stochastic gradient Langevin dynamics (SGLD). This algorithm samples
from a Bayesian posterior by adding artificial noise to the stochastic
gradient which, at long times, dominates the SGD noise. Also see
\citet{sato2014approximation} for a detailed convergence analysis of
the algorithm. Though elegant, one disadvantage of SGLD is that the
learning rate must be decreased to achieve the correct sampling
regime, and the algorithm can suffer from slow mixing times. Other
research suggests improvements to this issue, using Hamiltonian
Monte-Carlo~\citep{chen2014stochastic} or
thermostats~\citep{ding2014bayesian}. \citet{ma2015complete} give a
complete classification of possible stochastic gradient-based MCMC
schemes.

Above, we analyzed properties of stochastic gradient Fisher
scoring~\citep{ahn2012bayesian}. This algorithm speeds up mixing times
in SGLD by preconditioning a gradient with the inverse sampling noise
covariance. This allows constant learning rates, while maintaining
long-run samples from the posterior. In contrast, we do not aim
to sample exactly from the posterior. 
We describe how to tune the parameters of SGD such that its stationary distribution \emph{approximates} the posterior.

\citet{maclaurin2015early} also interpret SGD as a non-parametric
variational inference scheme, but with different goals and in a
different formalism. The paper proposes a way to track entropy changes
in the implicit variational objective, based on estimates of the
Hessian. As such, they mainly consider sampling distributions that are
not stationary, whereas we focus on constant learning rates and
distributions that have (approximately) converged. Note that their
notion of hyperparameters does not refer to model parameters but to parameters of SGD.

\parhead{Stochastic Optimization.} Stochastic gradient descent is an
active field~\citep{zhang2004solving,bottou1998online}. Many papers
discuss constant step-size SGD. \citet{bach2013non,
  flammarion2015averaging} discuss convergence rate of averaged
gradients with constant step size, while~\citet{defossez2015averaged}
analyze sampling distributions using quasi-martingale techniques.
\citet{toulis2014statistical} calculate the asymptotic variance of SGD
for the case of decreasing learning rates, assuming that the data is
distributed according to the model. None of these papers use variational arguments.

The fact that optimal preconditioning (using a decreasing
Robbins-Monro schedule) is achieved by choosing the inverse noise
covariance was first shown in~\citep{sakrison1965efficient}, but here
we derive the same result based on different arguments and
suggest a scalar prefactor. Note the optimal scalar learning rate of
$2/{\rm Tr}(BB^\top)$ can also be derived based on stability
arguments, as was done in the context of least mean square
filters~\citep{widrow1985adaptive}.

Finally,~\citet{chen2015bridging} also draw analogies between SGD and
scalable MCMC. They suggest annealing the posterior over iterations to
use scalable MCMC as a tool for global optimization. We follow the
opposite idea and suggest to use constant SGD as an approximate
sampler by choosing appropriate learning rate and preconditioners.

\parhead{Stochastic differential equations.} The idea of analyzing
stochastic gradient descent with stochastic differential equations is
well established in the stochastic approximation
literature~\citep{kushner2003stochastic, ljung2012stochastic}. Recent
work focuses on dynamical aspects of the
algorithm. \citet{li2015dynamics} discuss several one-dimensional
cases and momentum.~\citet{chen2015convergence} analyze
stochastic gradient MCMC and studied their convergence properties
using stochastic differential equations.

Our work makes use of the same formalism but has a different focus.
Instead of analyzing dynamical properties, we focus on stationary
distributions. Further, our paper introduces the idea of minimizing KL
divergence between multivariate sampling distributions and the
posterior.